\documentclass{article}
\usepackage{spconf,amsmath,graphicx}
\usepackage{xcolor}
\usepackage{amsmath,bbm}
\usepackage{amssymb}
\usepackage{booktabs}
\usepackage{multirow}

\title{DEEP FEATURE COMPRESSION FOR COLLABORATIVE OBJECT DETECTION}
\name{Hyomin Choi and Ivan V. Baji\'{c}}
\address{School of Engineering Science, Simon Fraser University, Burnaby, BC, Canada}
\begin{document}
%
\maketitle

\begin{abstract}
Recent studies have shown that the efficiency of deep neural networks in mobile applications can be significantly improved by distributing the computational workload between the mobile device and the cloud. This paradigm, termed \emph{collaborative intelligence}, involves communicating feature data between the mobile and the cloud. The efficiency of such approach can be further improved by lossy compression of feature data, which has not been examined to date. In this work we focus on collaborative object detection and study the impact of both near-lossless and lossy compression of feature data on its accuracy. We also propose a strategy for improving the accuracy under lossy feature compression. Experiments indicate that using this strategy, the communication overhead can be reduced by up to 70\% without sacrificing accuracy.   


\end{abstract}
\section{Introduction}
\label{sec:intro}
Mobile and Internet-of-Things (IoT)~\cite{xia2012internet} devices are increasingly relying on Artificial Intelligence (AI) engines to enable sophisticated applications such as personal digital assistants~\cite{PDA_SPM2017}, self-driving vehicles, autonomous drones, smart cities, and so on. The AI engines themselves are generally built on deep learning models. The most common way of deploying such models is to place them in the cloud and have the sensor data (images, speech, etc.) uploaded from the mobile to the cloud for processing. This is referred to as the \emph{cloud-only} approach. More recently, with smaller graphical processing units (GPUs) making their way into mobile/IoT devices, some deep models might be able to run on the mobile device, an approach referred to as \emph{mobile-only}.

A recent study~\cite{kang2017neurosurgeon} has examined a spectrum of possibilities in between the cloud-only and mobile-only extremes. Specifically, they considered splitting a deep network into two parts: the front end (consisting of an input layer and a number of subsequent layers), which runs on the mobile, and the back end (consisting of the remaining layers), which runs on the cloud. In this approach, termed \emph{collaborative intelligence}, the front end computes features up to some layer in the network, then these features are uploaded to the cloud for the remainder of the computation. The authors examined the energy consumption and latency associated with performing computation in this way, for various split points in typical deep models. Their findings indicate that significant savings can be achieved in both energy and latency if the network is split appropriately. They also proposed an algorithm called \emph{Neurosurgeon} to find the optimal split point, depending on whether energy or latency is to be minimized.

The reason why collaborative intelligence can be more efficient than cloud-only and mobile-only approaches is that the feature data volume in deep convolutional neural networks (CNNs) typically decreases as we move from the input to the output. Executing initial layers on the mobile will cost some energy and time, but if the network is split appropriately, we will end up with far less data to be uploaded to the cloud, which will save both transmission latency on the uplink and the energy used for radio transmission. Hence, on the balance, there may be a net benefit in energy and/or latency. Based on~\cite{kang2017neurosurgeon}, depending on the resources available (GPU or CPU on the mobile, speed and energy for wireless transmission, etc.), optimal split points for CNNs tend to be deep in the network. 

A recently released study~\cite{jointdnn} has extended the approach of ~\cite{kang2017neurosurgeon} to include model training and additional network architectures. While the network is again split between the mobile and the cloud, in the framework proposed in~\cite{jointdnn} the data can move both ways between the mobile and the cloud in order to optimize efficiency of both training and inference.

While~\cite{kang2017neurosurgeon,jointdnn} have established the potential benefits of collaborative intelligence, the issue of efficient transfer of feature data between the mobile and the cloud is largely unexplored. Specifically,~\cite{kang2017neurosurgeon} does not consider feature compression at all, while~\cite{jointdnn} uses 8-bit quantization of feature data followed by lossless compression, but does not examine the impact of such processing on the application. Feature compression can further improve the efficiency of collaborative intelligence by minimizing the latency and energy of feature data transfer. The impact of compressing the input has been studied in several CNN applications~\cite{dodge2016understanding, quality_model_for_od_using_compressed_video, hevc_for_object_detection} and the effects vary from case to case. However, the impact of feature compression has not been studied yet, to our knowledge.

In this work, we focus on a deep model for object detection and study the impact of feature compression on its accuracy. 
Section~\ref{sec:prev_work} presents preliminaries, while  Section~\ref{sec:proposed} describes the proposed methods. Experimental results and conclusions are presented in Sections~\ref{sec:experiments} and~\ref{sec:conclusion}, respectively.

\section{Preliminaries}
\label{sec:prev_work}
Object detection has been transformed in recent years with the advent of deep models that are able to simultaneously detect, localize, and classify objects in an image. Examples of such detectors include R-CNN~\cite{RCNN_obj}, SSD~\cite{SSD}, and YOLO~\cite{YOLO1}. This work focuses on YOLO. One of the major innovations of these detectors was that they were trained using a cost function composed of both bounding box error and object class error terms. The YOLO loss function is~\cite{YOLO1}:   
\begin{multline}
\lambda_{coord} \sum_{i=0}^{S^2} \sum_{j=0}^B \mathbbm{1}_{ij}^{obj} \left[(x_i - \hat{x_i})^2 + (y_i - \hat{y_i})^2 \right] \\
+ \lambda_{coord} \sum_{i=0}^{S^2} \sum_{j=0}^B \mathbbm{1}_{ij}^{obj} \left[(\sqrt{w_i} - \sqrt{\hat{w_i}})^2 + (\sqrt{h_i} - \sqrt{\hat{h_i}})^2\right]  \\
+\sum_{i=0}^{S^2} \sum_{j=0}^B \mathbbm{1}_{ij}^{obj} (C_i - \hat{C_i})^2  \\
+\lambda_{noobj} \sum_{i=0}^{S^2} \sum_{j=0}^B \mathbbm{1}_{ij}^{noobj} (C_i - \hat{C_i})^2 \\ 
+\sum_{i=0}^{S^2} \mathbbm{1}_{i}^{obj} \sum_{c \in classes} ({p_i}(c) - \hat{p_i}(c))^2
\label{eq:loss}
\end{multline}
where $(x_i,y_i)$ is the center of the ground truth bounding box, $w_i$ and $h_i$ are its width and height, $(\hat{x_i},\hat{y_i})$ is the center of the predicted bounding box whose width and height are $\hat{w_i}$ and $\hat{h_i}$, respectively. $C_i$ and $\hat{C_i}$ are the ground truth and predicted confidence scores corresponding to cell $i$, ${p_i}(c)$ and $\hat{p_i}(c)$ are the ground truth and predicted conditional probabilities for the object class $c$ in cell $i$, $\mathbbm{1}_{ij}^{obj}$ is equal to $1$ if the $j$-th bounding box in cell $i$ is responsible for prediction (i,e. box $j$ has the largest Intersection-over-Union among all boxes in cell $i$), and  $\mathbbm{1}_{ij}^{noobj}=1-\mathbbm{1}_{ij}^{obj}$. The scaling factors used are  $\lambda_{coord}=5$ and $\lambda_{noobj}=0.5$. 

\begin{figure}[t]
    \begin{minipage}[b]{1.0\linewidth}
    \centering
    \includegraphics[width=\textwidth]{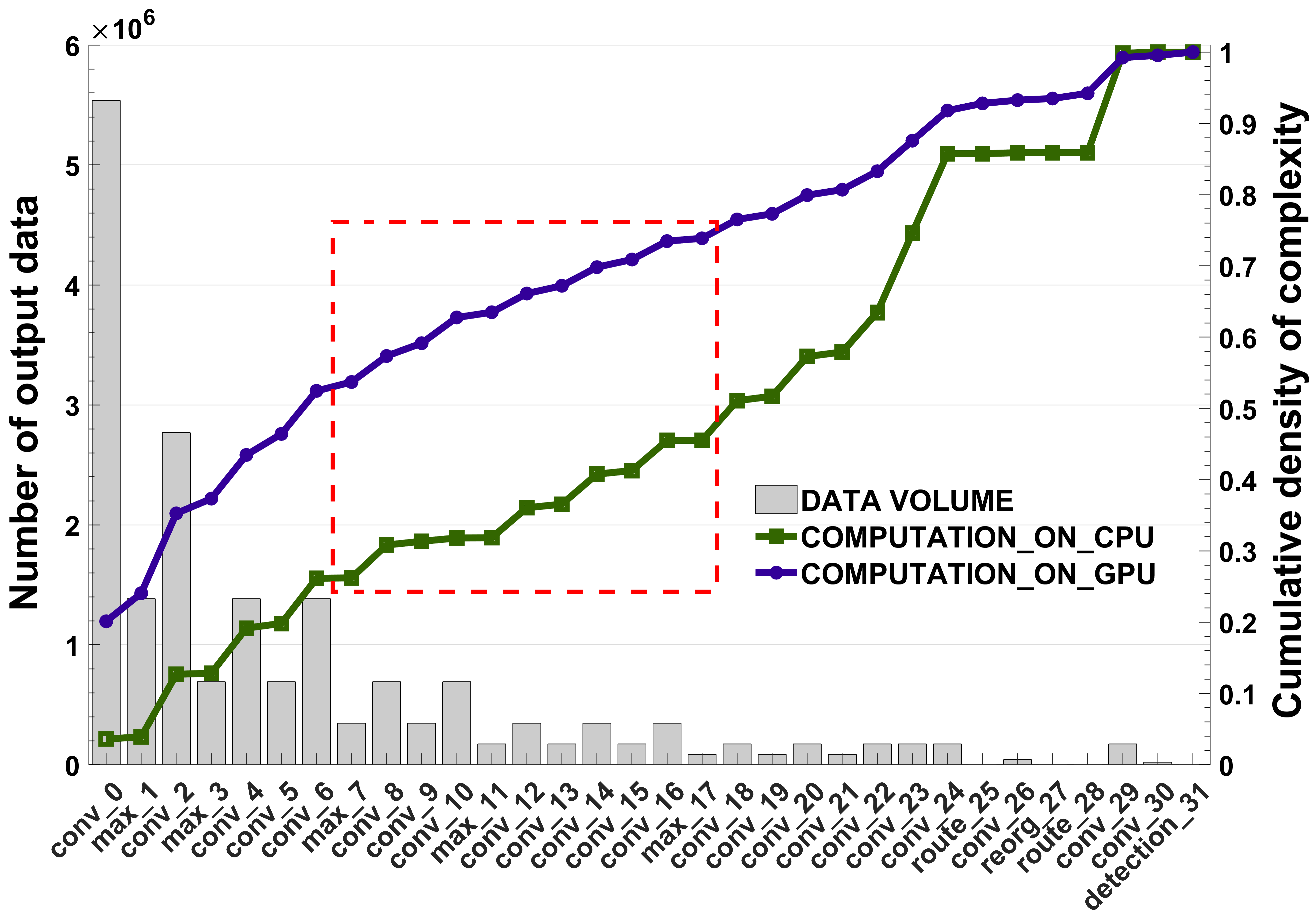}
    \end{minipage}
\vspace{-0.6cm}
\caption{Cumulative computation complexity and layer-wise output data volume}
\label{fig:complexity_and_volume}
\end{figure}

Our experiments in this work are based on the recent version of YOLO called YOLO9000~\cite{YOLO2}. Fig.~\ref{fig:complexity_and_volume} shows the feature data volume (number of feature samples) at the output of each layer of this model, as well as the cumulative computational cost (normalized execution time) as we move from the input layer towards the output. Computational cost was measured on a desktop machine with a Titan X GPU and Intel i7-6800K CPU over the images from a dataset described in Section~\ref{sec:experiments}. As seen in the figure, the feature data volume is fairly small starting with  max-pooling layer max\_7. Hence, this layer, or other downstream layers seem to be good points to split the network. Note that max-pooling (and other pooling) layers reduce the data volume, so from the point of view of data size, it is always advantageous to split the network at the output of the max-pooling layer rather than at its input. 

\begin{figure}[t]
    \begin{minipage}[b]{1.0\linewidth}
    \centering
    \includegraphics[width=\textwidth]{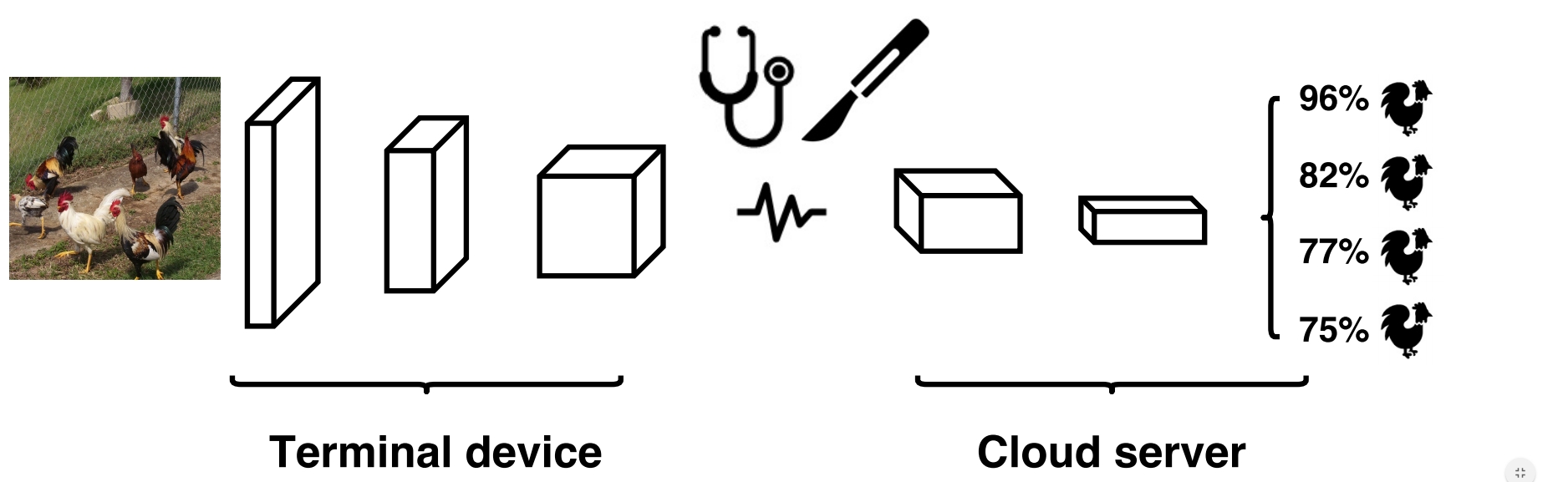}
    \centerline{(a)}\medskip
    \end{minipage}
    \hspace{-0.1cm}
    \begin{minipage}[b]{1.0\linewidth}
    \centering
    \includegraphics[width=\textwidth]{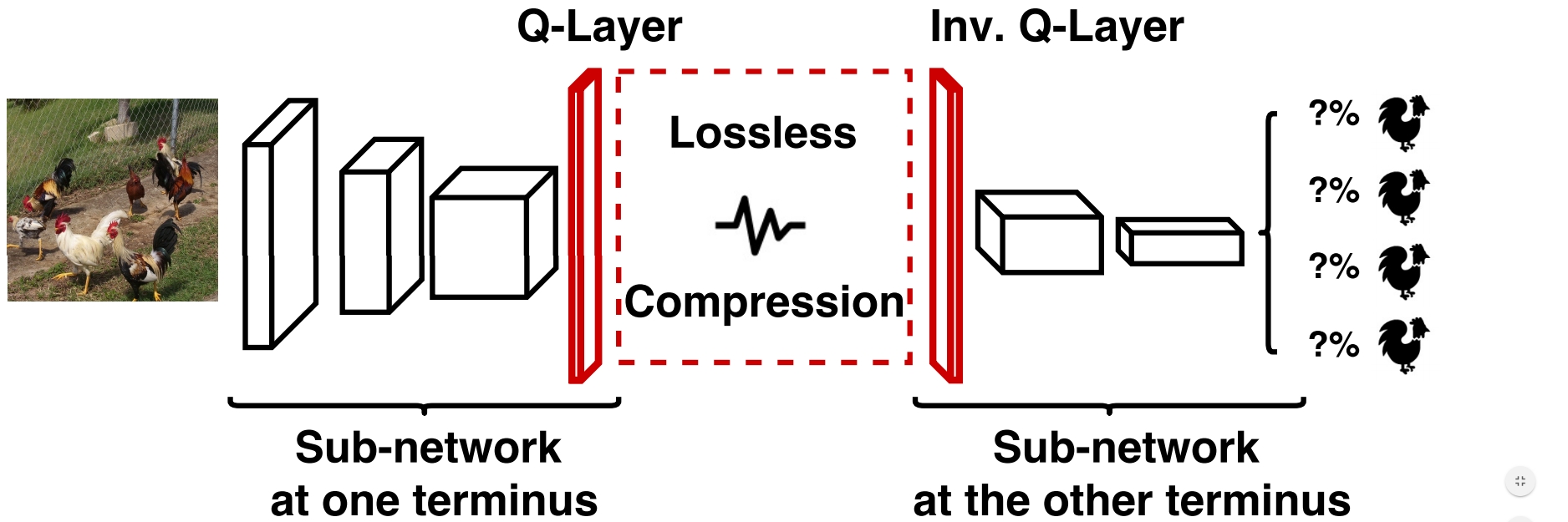}
    \centerline{(b)}\medskip
    \end{minipage}
    \hspace{-0.1cm}
    \begin{minipage}[b]{1.0\linewidth}
    \centering
    \includegraphics[width=\textwidth]{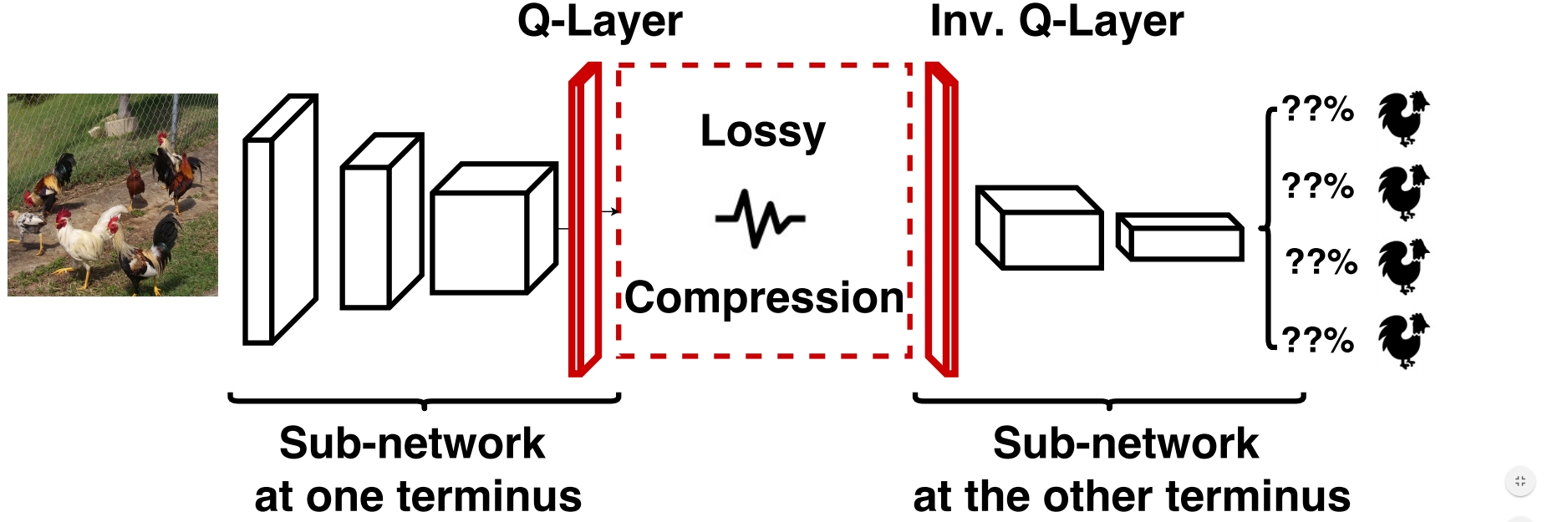}
    \centerline{(c)}\medskip
    \end{minipage}
    \hspace{-0.1cm}
\vspace{-0.7cm}
\caption{Three ways for mobile-cloud collaborative intelligence:  (a) Lossless transfer, (b) quantization followed by lossless compression, and (c) lossy compression.}
\label{fig:proposed_network}
\end{figure}

If we were to split the network at the output of some layer and transfer its feature data losslessly (as 32-bit floating point numbers) to the next layer (in the cloud), the accuracy would clearly stay the same as without the split\footnote{If the data is not corrupted during transmission, as assumed in~\cite{kang2017neurosurgeon,jointdnn}.}. This is the approach taken in~\cite{kang2017neurosurgeon}, and is illustrated in Fig.~\ref{fig:proposed_network}(a). But this is inefficient because the data likely contains some redundancy. 

A more efficient approach would be to compress the data prior to upload to the cloud. To achieve this, we could quantize the data, say to 8 bits per sample, then encode the quantized data losslessly. This is the approach taken in~\cite{jointdnn} with a lossless PNG encoder. It is illustrated in Fig.~\ref{fig:proposed_network}(b), where the quantization layer is called the Q-layer. This approach is \emph{near-lossless} because there is some quantization involved, and due to this quantization the accuracy of inference may be affected. An even more efficient approach to data transfer is to employ lossy compression after the Q-layer (Fig.~\ref{fig:proposed_network}(c)), but this will have an even greater impact on the accuracy. These issues are examined in Section~\ref{sec:experiments}.

\section{Proposed methods}
\label{sec:proposed}


\subsection{Quantization}
\label{ssec:qlayer}
In order to leverage existing codecs, the feature data is first quantized to 8-, 10-, or 12-bit precision in a Q-layer, which is inserted at the split point. Let $\mathbf{V} \in \mathbb{R}^{N \times M \times C}$ be the tensor containing the feature data at the point of split, with $N$ rows, $M$ columns, and $C$ channels. Let $\textup{min}(\mathbf{V})$ and $\textup{max}(\mathbf{V})$ be the minimum and maximum value in $\mathbf{V}$, respectively. Quantization with $n_{bit}$-precision and the corresponding inverse quantization in the inverse Q-layer are performed as  
\begin{equation} 
\widetilde{\mathbf{V}} = \textup{round} \left (\frac{\mathbf{V}-\textup{min}(\mathbf{V})}{\textup{max}(\mathbf{V})-\textup{min}(\mathbf{V})}\cdot (2^{n_{bit}}-1)\right )
\label{eq:quantized_vector}
\end{equation}

\begin{equation} 
\widehat{\mathbf{V}} = \frac{\widetilde{\mathbf{V}}\cdot (\textup{max}(\mathbf{V})-\textup{min}(\mathbf{V}))}{2^{n_{bit}}-1}+\textup{min}(\mathbf{V})
\label{eq:dequantized_vector}
\end{equation}

\noindent where $\widetilde{\mathbf{V}}$ is the quantized feature tensor, $\widehat{\mathbf{V}}$ is the de-quantized feature tensor, and $\textup{round}(\cdot)$ represents rounding to the nearest integer. $\textup{min}(\mathbf{V})$ and $\textup{max}(\mathbf{V})$ need to be stored as 32-bit floats (8 Bytes total) and transferred to the cloud for de-quantization. This is taken into account when computing total bits in the experiments. Note that in some cases, such as when the previous activation layer is sigmoid or ReLU (assumed in~\cite{jointdnn}), we can consider $\textup{min}(\mathbf{V})=0$ and avoid transmitting it, but for more general activation layers such as Leaky ReLU (which is used in YOLO9000) this parameter is required.     

\vspace{-5pt}

\subsection{Compression}
\label{ssec:tiling}

\begin{figure}[t]
    \begin{minipage}[b]{0.48\linewidth}
    \centering
    \includegraphics[width=\textwidth]{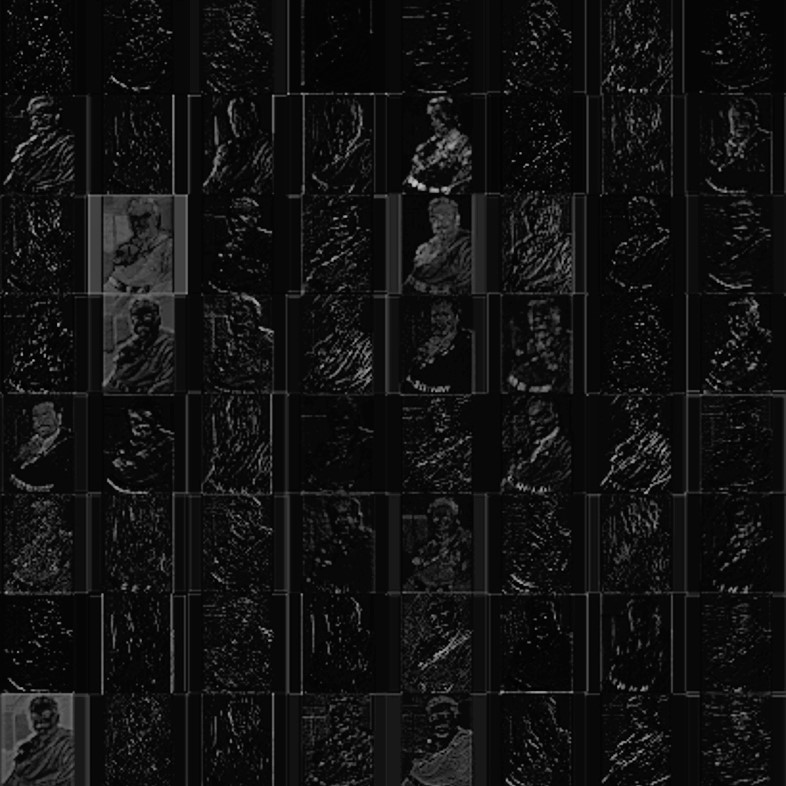}
    \centerline{(a)}\medskip
    \label{fig:tiling}
    \end{minipage}
    \hspace{0.04cm}
    \begin{minipage}[b]{0.48\linewidth}
    \centering
    \includegraphics[width=\textwidth]{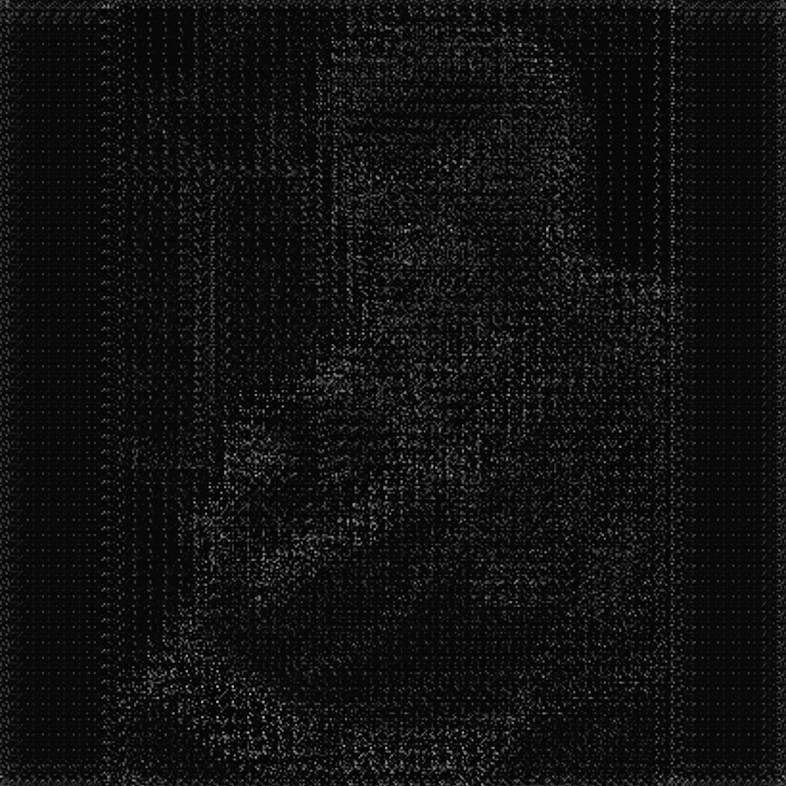}
    \centerline{(b)}\medskip
    \label{fig:quilting}
    \end{minipage}
\vspace{-0.5cm}
\caption{Combining feature channels into an image by (a) tiling and (b) quilting.}
\label{fig:tiled_map}
\end{figure}

Quantized feature tensor $\widetilde{\mathbf{V}}$ can be encoded by a number of existing codecs. If we interpret the $N \times M \times C$ tensor as $C$ frames of size $N\times M$, we could employ a video codec to compress it. We could combine groups of feature channels into larger frames, to end up with less than $C$ frames with larger resolution. Finally, all channels could be combined into a single image. Even in this case there are a number of possibilities, such as tiling (Fig.~\ref{fig:tiled_map}(a)), where the entire channel is placed in the image as a tile, followed by another tile, and so on, and quilting (Fig.~\ref{fig:tiled_map}(b)), where neighboring samples come from different channels. We tested a number of such methods and found that simple tiling by channel index provided the best results, so we use that method from here on. Hence, tiled feature channels are compressed as a still image.

For compression, we employ high efficiency video coding (HEVC)~\cite{hevc} standard, specifically HEVC Range extension (RExt)~\cite{flynn2016overview} which supports 4:0:0 sample format with various bit-depths. HM16.12~\cite{HM16.12} in the experiments and all coding tools and configuration follow common test condition~\cite{hevc_ctc}. RDOQ tool is turned off and the coding tree unit (CTU) size is set to 16$\times$16, because the feature channel resolution is relatively small deep in the network.

\vspace{-5pt}

\subsection{Compression-augmented training}
\label{ssec:training}

As will be seen in Section~\ref{sec:experiments}, Q-layer quantization followed by lossless compression has little effect on the accuracy. However, lossy compression may affect the accuracy, especially when the quantization parameter (QP) is high. This loss in accuracy can be somewhat compensated by compression-augmented training. Instead of using the network parameters (weights) supplied with the model, we re-train the model by considering lossy compression at the point of split. During training, at each forward pass through the network, the feature data at the split point is tiled and compressed using a randomly chosen QP value. In our experiments we used  QP in the range [Lossless, 22, 27, 32, 37]. After compression, the decompressed data is passed further down the network. 

This kind of compression augmentation can be interpreted as a form of regularization, where quantization noise is inserted into an intermediate layer deep in the network. It encourages the network to learn the downstream weights (from the split point) that provide good accuracy when processing decompressed features, and also to learn upstream weights that generate features that are robust to compression.

\section{Experiments}
\label{sec:experiments}


Following~\cite{YOLO2}, a total of 16,551 images from VOC2007 and VOC2012 datasets~\cite{pascal-voc-2007,pascal-voc-2012} are used for training and another 4,952 images from VOC2007 for testing. Twenty different object classes are represented in the dataset. 



\begin{figure}[t]
    \begin{minipage}[b]{0.95\linewidth}
    \centering
    \includegraphics[width=\textwidth]{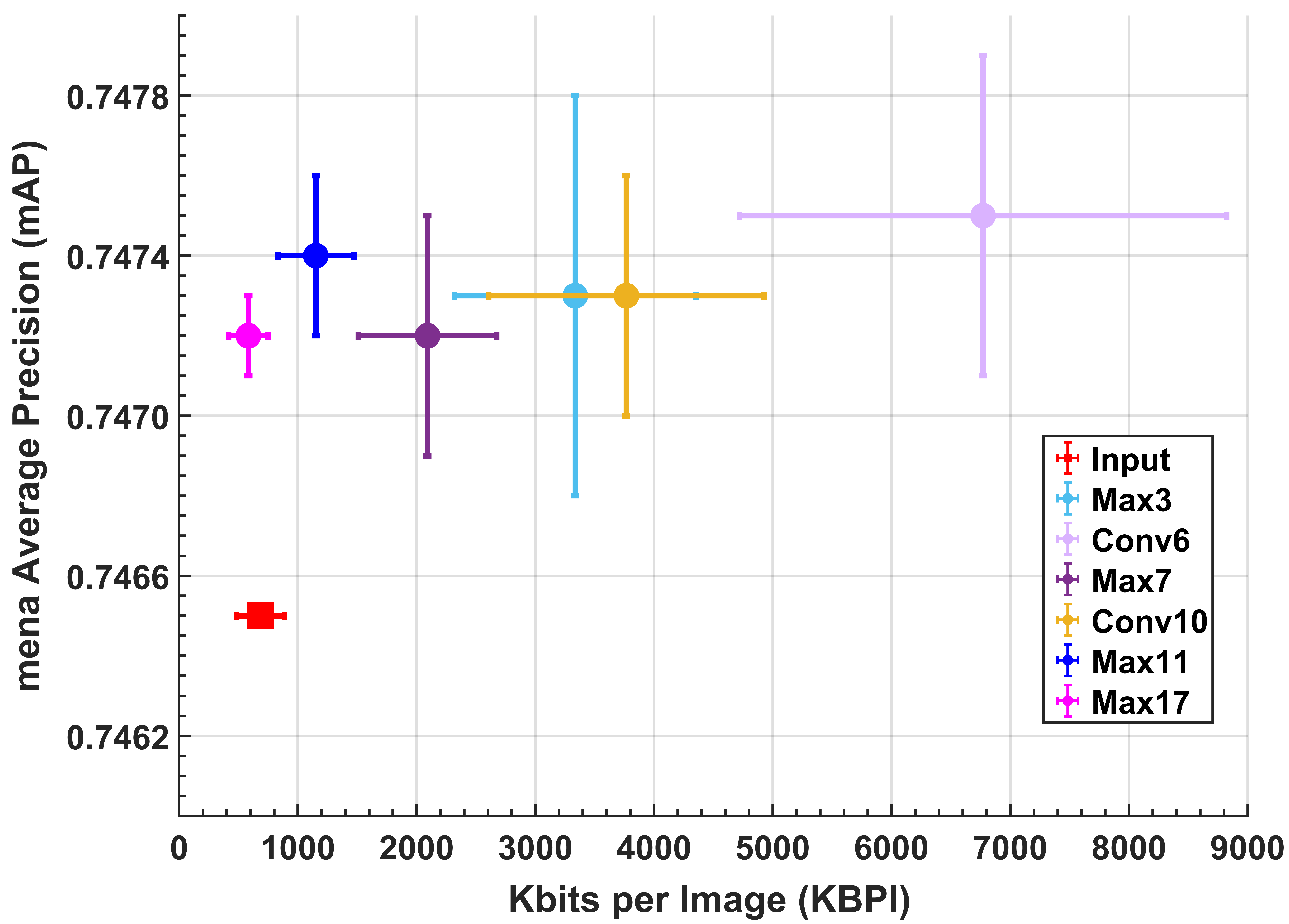}
    \end{minipage}
    \hspace{0.04cm}
\vspace{-0.1cm}
\caption{mAP vs. KBPI for lossless deep feature compression}
\label{fig:lossless_comp}
\vspace{-0.1cm}
\end{figure}

We first test the impact of lossless compression (after the Q-layer) on accuracy. As is common with multi-class object detectors~\cite{VOC}, we use mean Average Precision (mAP) as a measure of accuracy, and look at its variation with 8-bit, 10-bit and 12-bit quantization in the Q-layer. The compression of feature data is quantified using average Kbits per image (KBPI). Fig.~\ref{fig:lossless_comp} presents mAP versus KBPI for various split points in the network. Vertical bars show the standard deviation of mAP at a given average KBPI, while horizontal bars show the standard deviation of KBPI for the corresponding average mAP. The red square indicates the operating point achieved by the cloud-only approach, without network splitting and uploading the input JPEG images to the cloud. 

As seen in the figure, when the split point is close to the input (e.g. max\_3, conv\_6 or conv\_10 layers), the data volume is too large, and even with lossless compression of feature data, it is more efficient to simply upload input images to the cloud. But as we move down the network, it becomes more advantageous to upload  feature data. Meanwhile, the mAP does not change much - scores around 0.7465-0.7475 are achieved for all the cases. Hence, lossless compression of deep features (following 8-, 10-, or 12-bit quantization) has only a minor influence on accuracy, but also provides limited (if any) bit savings for data transfer to the cloud.

\begin{figure}[t]
    \begin{minipage}[b]{0.95\linewidth}
    \centering
    \includegraphics[width=\textwidth]{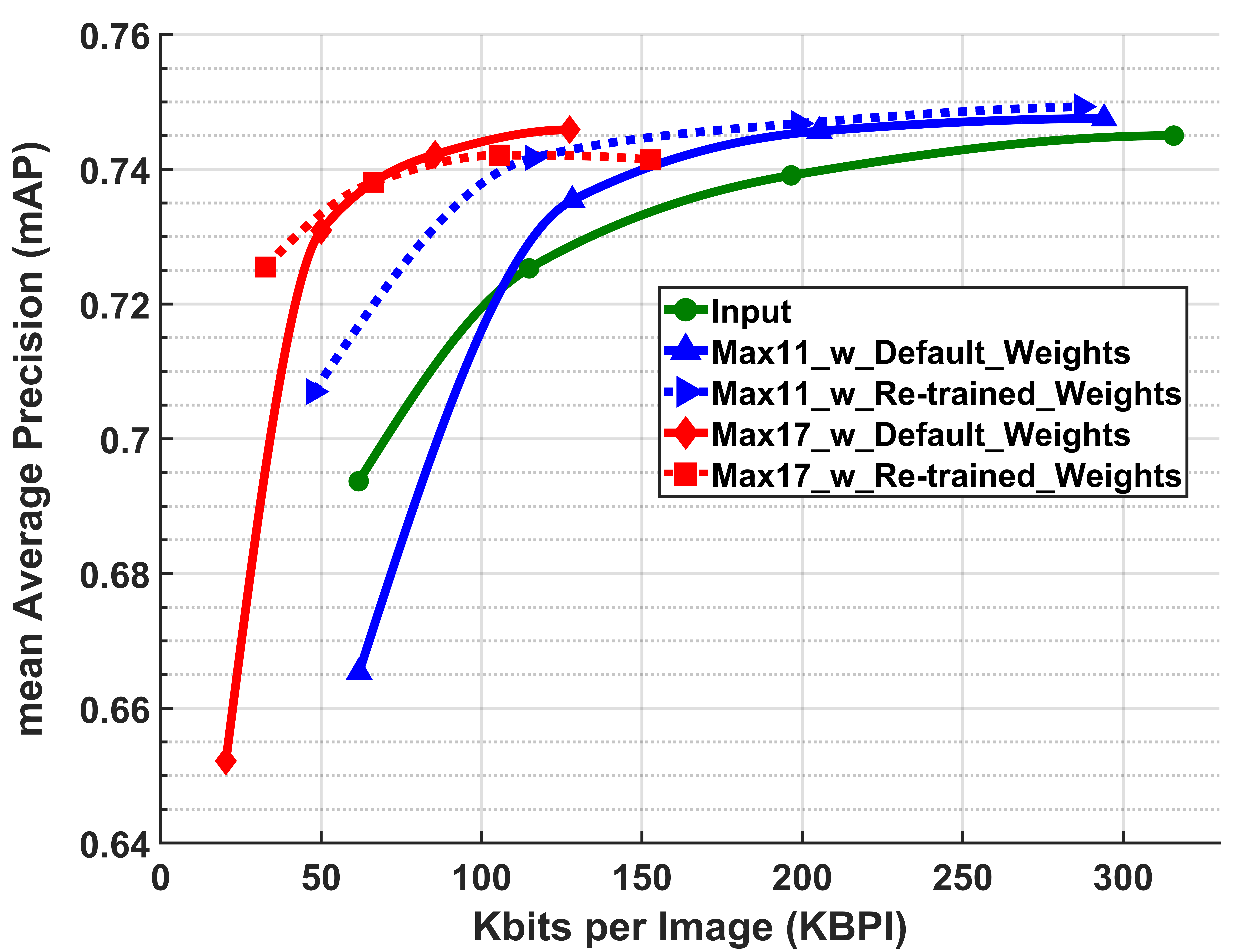}
    \end{minipage}
\vspace{-0.2cm}
\caption{mAP vs. KBPI for lossy deep feature compression}
\label{fig:rate_map_curve}
\vspace{-0.1cm}
\end{figure}
Lossy compression offers significant bit savings, but care must be taken to minimize the loss of accuracy. In order to evaluate the impact of lossy compression, we show mAP vs. KBPI curves in Fig.~\ref{fig:rate_map_curve}. The green curve corresponds to compressing the input image, as the default cloud-only approach. The blue curves correspond to splitting the network at the output of max\_11 layer, and red curves correspond to the split after the max\_17 layer. In each case, the solid line corresponds to using default YOLO9000 weights while the dashed line corresponds to using the weights obtained by compression-augmented training, starting from the pre-trained weight, ``Darknet19 448x448'', for ImageNet classification~\cite{ILSVRC15} and following the training procedure in ~\cite{darknet}. As seen in the figure, lossy compression can provide significant bit savings over the cloud-only approach, while compression-augmented training further extends the range of useful compression levels for a given mAP.




To quantify the differences between various cases, we adopt a Bjontegaard Delta (BD) approach~\cite{Bjontegaard}. Specifically, we use the BD calculation to compute BD-KBPI-mAP, which indicates the average difference in KBPI at the same mAP. The results are shown in Table~\ref{tbl:performance}, where the default case against which the comparison is made is the cloud-only approach.
As shown in the table, compressing features at the output of max\_11 (max\_17) while using default weights would give an average saving of 6\% (60\%) at the same mAP compared to cloud-only approach. Meanwhile, the weights obtained through compression-augmented training would provide an additional bit saving of 39\% (10\%), for the total of up to 45\% (70\%) bit savings.



\begin{table}[b]
\renewcommand{\arraystretch}{0.85}
\centering
\vspace{-0.3cm}
\caption{BD-KBPI-mAP of lossy feature compression vs. cloud-only approach}
\label{tbl:performance}
\scalebox{0.9}{
\smallskip\noindent
\resizebox{\linewidth}{!}{
\begin{tabular}{ccc}
\toprule
Split at  &  Default weights  &  Re-trained weights\\
\midrule
max\_11  & $-6.09\%$ & $\mathbf{-45.23\%}$ \\
max\_17  & $-60.30\%$ & $\mathbf{-70.30\%}$ \\
\bottomrule

\end{tabular}}}
\end{table}



\section{Conclusions}
\label{sec:conclusion}

We studied deep feature compression for collaborative object detection between the mobile and the cloud. We examined the impact of compression on detection accuracy and showed that lossless compression of 8-bit (or higher) quantized data does not have much impact on the accuracy. Lossy compression provides higher bit savings, but also affects the accuracy. To compensate for this, we proposed compression-augmented training, which is able to extend the range of useful compression levels for a desired accuracy.


\bibliographystyle{IEEEbib}
\bibliography{ref}

\begin{thebibliography}{10}

\bibitem{xia2012internet}
F.~Xia, L.~T. Yang, L.~Wang, and A.~Vinel,
\newblock ``Internet of things,''
\newblock {\em Int. Journal of Communication Systems}, vol. 25, no. 9, pp.
  1101, 2012.

\bibitem{PDA_SPM2017}
R.~Sarikaya,
\newblock ``The technology behind personal digital assistants: An overview of
  the system architecture and key components,''
\newblock {\em IEEE Signal Processing Magazine}, vol. 34, no. 1, pp. 67--81,
  Jan. 2017.

\bibitem{kang2017neurosurgeon}
Y.~Kang, J.~Hauswald, C.~Gao, A.~Rovinski, T.~Mudge, J.~Mars, and L.~Tang,
\newblock ``Neurosurgeon: Collaborative intelligence between the cloud and
  mobile edge,''
\newblock in {\em Proc. 22nd ACM Int. Conf. Architectural Support for
  Programming Languages and Operating Systems}, 2017, pp. 615--629.

\bibitem{jointdnn}
A.~E. Eshratifar, M.~S. Abrishami, and M.~Pedram,
\newblock ``{JointDNN}: an efficient training and inference engine for
  intelligent mobile cloud computing services,''
\newblock {\em arXiv preprint arXiv:1801.08618}, 2018.

\bibitem{dodge2016understanding}
S.~Dodge and L.~Karam,
\newblock ``Understanding how image quality affects deep neural networks,''
\newblock in {\em IEEE Int. Conf. Quality of Multimedia Experience (QoMEX'16)}.
  IEEE, 2016, pp. 1--6.

\bibitem{quality_model_for_od_using_compressed_video}
L.~Kong, R.~Dai, and Y.~Zhang,
\newblock ``A new quality model for object detection using compressed videos,''
\newblock in {\em Proc. IEEE ICIP'16}, Sep. 2016.

\bibitem{hevc_for_object_detection}
H.~Choi and I.~V. Baji\'{c},
\newblock ``High efficiency compression for object detection,''
\newblock in {\em Proc. IEEE ICASSP'18}, Apr. 2018,
\newblock to appear.

\bibitem{RCNN_obj}
R.~Girshick, J.~Donahue, T.~Darrell, and J.~Malik,
\newblock ``Rich feature hierarchies for accurate object detection and semantic
  segmentation,''
\newblock in {\em Proc. IEEE CVPR'14}, 2014, pp. 580--587.

\bibitem{SSD}
W.~Liu, D.~Anguelov, D.~Erhan, C.~Szegedy, S.~Reed, C.-Y. Fu, and A.~C. Berg,
\newblock ``{SSD:} single shot multibox detector,''
\newblock in {\em Proc. ECCV}, 2016.

\bibitem{YOLO1}
J.~Redmon, S.~Divvala, R.~Girshick, and A.~Farhadi,
\newblock ``You only look once: Unified, real-time object detection,''
\newblock in {\em Proc. IEEE CVPR'16}, Jun. 2016, pp. 779--788.

\bibitem{YOLO2}
J.~Redmon and A.~Farhadi,
\newblock ``{YOLO9000:} better, faster, stronger,''
\newblock in {\em Proc. IEEE CVPR'17}, Jul. 2017, pp. 6517--6525.

\bibitem{hevc}
G.~J. Sullivan, J.-R. Ohm, W.-J. Han, and T.~Wiegand,
\newblock ``Overview of the high efficiency video coding {(HEVC)} standard,''
\newblock {\em IEEE Trans. Circuits Syst. Video Technol.}, vol. 22, no. 12, pp.
  1649--1668, 2012.

\bibitem{flynn2016overview}
D.~Flynn, D.~Marpe, M.~Naccari, T.~Nguyen, C.~Rosewarne, K.~Sharman, J.~Sole,
  and J.~Xu,
\newblock ``Overview of the range extensions for the hevc standard: Tools,
  profiles, and performance,''
\newblock {\em IEEE Trans. Circuits Syst. Video Technol.}, vol. 26, no. 1, pp.
  4--19, 2016.

\bibitem{HM16.12}
``{HEVC} reference software ({HM} 16.12),''
  {https://hevc.hhi.fraunhofer.de/trac/hevc/browser/tags/HM-16.12},
\newblock Accessed: 2017-05-27.

\bibitem{hevc_ctc}
F.~Bossen,
\newblock ``Common {HM} test conditions and software reference
  configurations,''
\newblock in {\em ISO/IEC JTC1/SC29 WG11 {m28412}, {JCTVC-L1100}}, Jan. 2013.

\bibitem{pascal-voc-2007}
M.~Everingham, L.~Van~Gool, C.~K.~I. Williams, J.~Winn, and A.~Zisserman,
\newblock ``The {PASCAL} {V}isual {O}bject {C}lasses {C}hallenge 2007
  {(VOC2007)} {R}esults,'' http://host.robots.ox.ac.uk/pascal/VOC/voc2007/.

\bibitem{pascal-voc-2012}
M.~Everingham, L.~Van~Gool, C.~K.~I. Williams, J.~Winn, and A.~Zisserman,
\newblock ``The {PASCAL} {V}isual {O}bject {C}lasses {C}hallenge 2012
  {(VOC2012)} {R}esults,'' http://host.robots.ox.ac.uk/pascal/VOC/voc2012/.

\bibitem{VOC}
M.~Everingham, L.~van Gool, C.~K.~I. Williams, J.~Winn, and A.~Zisserman,
\newblock ``The {PASCAL} visual object classes (voc) challenge,''
\newblock {\em Int. Journal of Computer Vision}, vol. 88, no. 2, pp. 330--338,
  2010.

\bibitem{ILSVRC15}
O.~Russakovsky, J.~Deng, H.~Su, J.~Krause, S.~Satheesh, S.~Ma, Z.~Huang,
  A.~Karpathy, A.~Khosla, M.~Bernstein, A.~C. Berg, and Li~F,
\newblock ``{ImageNet Large Scale Visual Recognition Challenge},''
\newblock {\em Int. Journal of Computer Vision}, vol. 115, no. 3, pp. 211--252,
  2015.

\bibitem{darknet}
J.~Redmon,
\newblock ``{Darknet: Open source neural networks in C.},''
  http://pjreddie.com/darknet/, 2013-2017,
\newblock Accessed: 2017-10-19.

\bibitem{Bjontegaard}
G.~Bjontegaard,
\newblock ``Calculation of average {PSNR} differences between {RD}-curves,''
  Apr. 2001,
\newblock VCEG-M33.

\end{thebibliography}

\end{document}